%% file: main.tex
\documentclass[letterpaper]{article} 
\usepackage{aaai2027}  
\usepackage[hyphens]{url}  
\usepackage{graphicx} 
\usepackage{natbib}  
\usepackage{caption} 
\usepackage{algorithm}
\usepackage{algpseudocode}
\usepackage{amssymb}

\usepackage{amsmath}
\usepackage{booktabs}
\usepackage{multirow}
\usepackage{xcolor}
\usepackage{diagbox}

\title{Label-Free Role-Aware Segmentation of Classroom Recordings:\\
Recovering the Teacher from Transferable Structure}
\author{
    Nitin Jain\textsuperscript{\rm 1},
    Tim Fütterer\textsuperscript{\rm 1},
    Babette Bühler\textsuperscript{\rm 2},
    Çağatay Yıldız\textsuperscript{\rm 1,\rm 3}
}
\affiliations{
    \textsuperscript{\rm 1}Hector Research Institute of Education Sciences and Psychology, University of Tübingen\\
    \textsuperscript{\rm 2}Chair of Human-Centered Technologies for Learning, Technical University of Munich\\
    \textsuperscript{\rm 3}Tübingen AI Center, University of Tübingen\\
    nitin96.jain@gmail.com,~babette.buehler@tum.de,~ \{tim.fuetterer, cagatay.yildiz\}@uni-tuebingen.de
}

\input{tables.tex}  

\begin{document}
\maketitle

\begin{abstract}
How much the teacher talks and when are among the most informative and costly-to-measure signals of teaching. Yet, classroom audio is among the most sensitive data an AI system can process: recordings capture minors and fall under GDPR, preventing raw audio from leaving the recording device. We present a teacher-student segmentation pipeline designed to meet exactly this constraint: label-free at inference, running on a recording device, and requiring no per-session calibration. The pipeline chains a lightweight voice activity detector, compactly pretrained speaker embeddings, agglomerative clustering, and a small logistic regression that assigns the teacher role based on session-relative structure rather than voice identity. We show that our structural cue transfers across classrooms, languages, and acoustic conditions.

In leave-one-teacher-out cross-validation on 292 recordings from 63 teachers across English and German corpora, our pipeline achieves teacher F1 of 0.81 and 0.87, respectively, outperforming both enrollment-based baselines and a 30-second supervised enrollment ceiling.
On a commodity smartphone, its performance matches pyannote, a strong neural diarization tool, while running 4.8$\times$ faster at comparable peak memory at a real-time factor of {$\sim$0.18}. 
We also find that transcribing filtered audio only marginally increases teacher's word error rate from $0.26$ to $0.28$.
Residual failures concentrate where teacher and student voices are not acoustically separable.
However, we show how to flag these sessions label-free at inference so that they can be routed to human review, enabling selective, privacy-preserving deployment. 
\end{abstract}

\section{Introduction}

How much a teacher talks, and when, provides informative but costly-to-measure signals of teaching \citep{TAO2024100638}. Distinguishing teacher from student speech enables analyses of talk-time balance, turn-taking, and instructional dialogue, but still requires intensive human annotation \citep{wang2025jedm, kelly2018authenticity, donnelly2017words}. Automating this distinction supports scalable feedback and downstream analyses of classroom discourse and student engagement \citep{schlotterbeck2021tarta, hou2025multimodal}.

Speaker diarization identifies \emph{who spoke when} \citep{anguera2012review, park2022review}, but classroom deployment adds two requirements. First, a deployed system meets a new, unknown teacher in every session, and it must identify the teacher in unseen classrooms without enrollment audio or speaker-specific fine-tuning. Second, recordings are among the most sensitive data because they involve minors and sensitive interactions. In our intended downstream workflow, teacher speech must leave the device for further processing, making it necessary to identify and filter teacher segments locally before transmission. This motivates local processing without transferring raw audio or retaining reference voiceprints \citep{sumer2020anonymisation, granqvist2020federated}. Together, these constraints rule out the commonly used large, cloud-based GPU neural diarization pipelines with enrollment and demand a label-free, CPU-based inference system.

We present such a system. Our pipeline combines lightweight voice activity detection \citep{silero2024}, compact pretrained speaker embeddings \citep{wang2023wespeaker}, and agglomerative clustering with a small logistic regression that assigns teacher/student roles using session-relative features. The strongest cue we consider is \texttt{centrality}: the cosine similarity between speech segments in a lesson and a lesson's mean embedding. When the teacher is a recurrent, acoustically coherent voice in a classroom, this mean provides an in-session reference for recovering the teacher across fragmented segments. The role classifier is trained on annotated recordings; inference uses only audio and the frozen models.

We evaluate on 292 recordings from 63 teachers: 151 close-microphone English lessons and 141 far-field German lessons. Leave-one-teacher-out teacher F1 reaches $0.81$ and $0.87$, respectively, exceeding pseudo-enrollment and supervised enrollment baselines and remaining competitive with pyannote \citep{plaquet2023powerset, bredin2023pyannote}. Training on either language transfers to the other with no loss or target-language supervision. On Samsung Galaxy S24+ smartphone, the pipeline runs at real-time factor $\approx 0.18$, $4.8\times$ faster than the matched pyannote reference at comparable memory. We additionally show minimal degradation in a downstream transcription task, audit ground-truth transcript-derived labels, and evaluate label-free diagnostics for routing unreliable sessions to human review.

\paragraph{Task and constraints.} We assign every instant of a single-channel recording one of three labels: \texttt{silence}, \texttt{teacher}, or \texttt{student}, grouping all students into a single role rather than distinguishing individuals. Labels are used only to train and calibrate the classifier; inference requires neither enrollment audio nor per-session tuning. The overall system is to be deployed on smartphones.

\begin{figure}[t]
    \centering
    \includegraphics[width=0.99\linewidth]{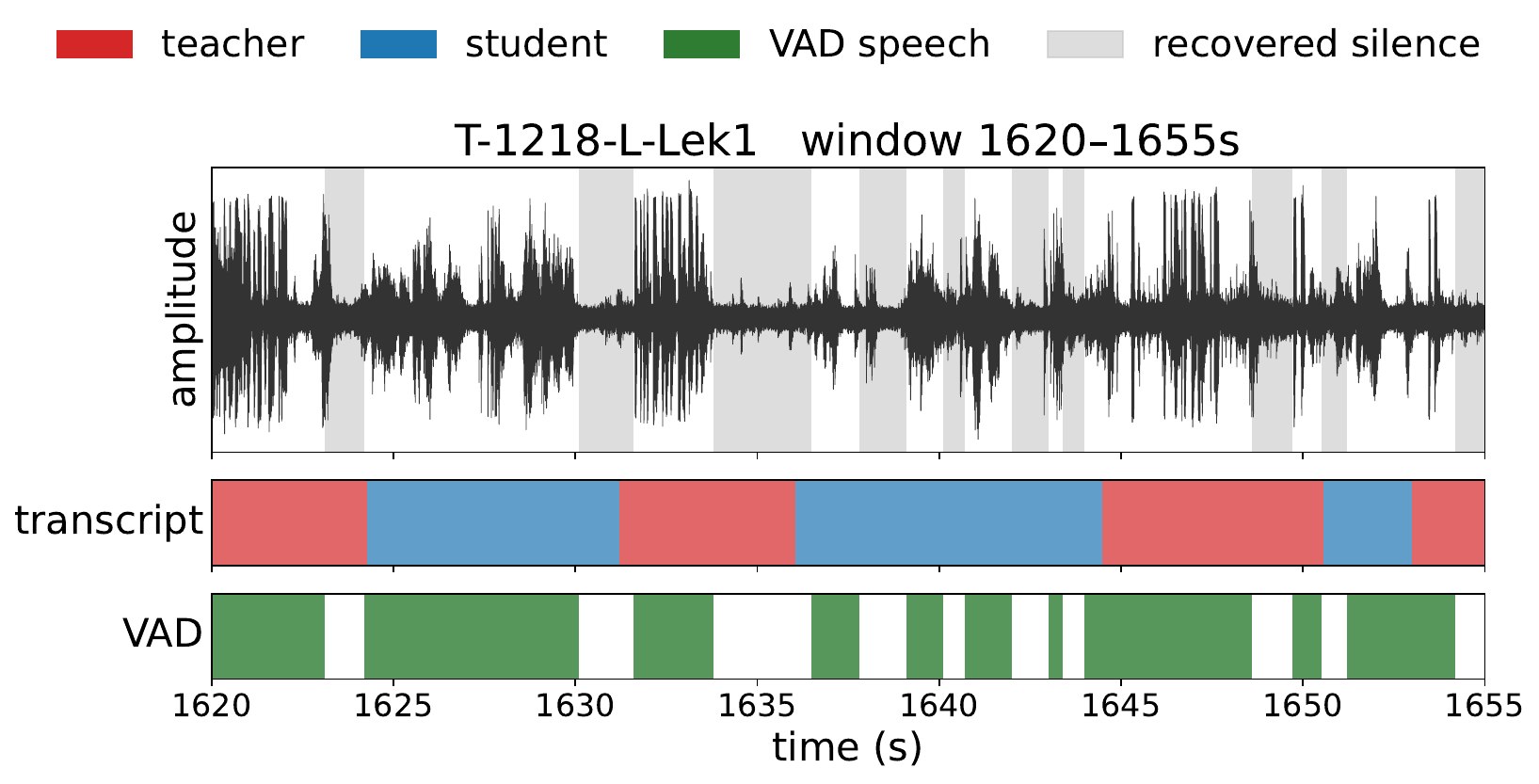}
    \vspace{-.25cm}
    \caption{35\,s segment of a classroom recording, ground truth transcript annotation (who spoke), and the output of the VAD stage of our pipeline, identifying silent moments.}
    \label{fig:vad}
\end{figure}

\section{Datasets and problems with their labels}
\label{sec:data}
We use two corpora in two acoustic regimes:

\begin{itemize}
  \item \textbf{Virginia (English, \citet{pianta2016ncrece}).} Classroom videos in which teachers wear a microphone. 151 lessons have human transcripts (24 distinct teachers).
  \item \textbf{GTI (German, Global Teaching InSights \citet{oecd2020gti})} Room-microphone recordings, 141 transcribed lessons and tutoring sessions (39 distinct teachers). This dataset was initially referred to as the Teaching and Learning International Survey (TALIS) Video Study.
\end{itemize}

\textbf{Onset-anchored labels.~~}
Transcript timestamps mark turn onsets, not offsets. For instance, a span labeled \texttt{Teacher 38\,$\rightarrow$\,56} ends only where the next speaker begins, and hides pauses, backchannels, and interjections that never received their own onset. VAD confirms that only $69\%$ (English) and $46\%$ (German) of teacher-labeled interval time is actually voiced. Treating entire intervals as teacher speech therefore contaminates training labels and distorts evaluation. At \emph{training} we intersect transcript intervals with VAD (Section~\ref{sec:vad}); at \emph{evaluation} we trust only the first $4$\,s after each onset (Section~\ref{sec:metric}). The strategy ranking remains stable across alternative window lengths (Appendix~\ref{app:headsweep}).


\textbf{Boundary timing noise.~~}
Onset timestamps are recorded to whole-second resolution, so the final $\sim$$1$\,s before each speaker change is still credited to the outgoing speaker although the incoming one has usually begun. Excluding that second from the ground truth raises teacher F1 from $0.85$ to $0.87$. Because the model is already correct on these frames and only the label is wrong, we do not revise our numbers upward and treat the reported F1 as a conservative floor, with roughly two points of residual error attributable to transition-zone label noise rather than model error.

\textbf{Human audit.~~}
We sampled $5{,}000$ clips across six groups: two where our pipeline and the transcript disagree on the speaker's role, two where they agree on teacher or student, and two containing VAD-designated silence. A trained annotator, blind to both pipeline outputs and transcripts, labeled the clips. Duration-weighted estimates indicate substantial transcript error: $63\%$ of sampled duration labeled teacher by our pipeline but student by the transcript was judged to be teacher speech. Corrections almost entirely improve teacher precision without degrading recall. Appendix~\ref{app:gold} details the sampling, annotation, and corrected metrics.

\section{Method}
\label{sec:method}

\begin{algorithm}[t]
  \caption{Assign teacher/student roles to speaker clusters}
  \label{alg:classifier}
  \begin{algorithmic}[1]
  \Require A classroom audio recording
  \Ensure  Each cluster labelled \texttt{teacher} or \texttt{student}

  \medskip
  \State \textbf{Step 1: Embed and cluster.}
         Run VAD to filter out the silent regions. Then, split the active regions into 1.25\,s speech windows, embed each window with WeSpeaker \texttt{english\_resnet34}, and group the window embeddings into speaker clusters by agglomerative clustering.

  \medskip
  \State \textbf{Step 2: Describe each cluster.}
         For every cluster, compute eight structural features from the audio (no labels). \texttt{centrality} is defined in Sec.~\ref{sec:features}; the rest in
         App.~\ref{app:features}.

  \medskip
  \State \textbf{Step 3: Score each cluster.}
         Pass these features through a logistic regression model trained to predict if a summary is student or teacher. The model returns a \emph{teacher probability} $\hat{p}_c \in [0,1]$ for each cluster $c$.

\medskip
\State \textbf{Step 4: Assign teacher role.} Label as \texttt{teacher} all clusters with $\hat{p}_c \geq \tau$. Label all remaining clusters \texttt{student}. We find that the optimal $\tau=0.45$

  \end{algorithmic}
  \end{algorithm}

The inference pipeline is four stages: voice activity detection (VAD), embedding, clustering, and teacher selection.

\subsection{Voice activity detection and why we need it}
\label{sec:vad}

Silero VAD \citep{silero2024} identifies speech regions, assigning the \texttt{silence} label and restricting embedding extraction at \emph{inference}. During \emph{training}, we intersect detected speech with transcript intervals and assign each retained sub-span the interval's role label.

A human audit, stratified by frame energy in VAD-designated silence, estimates speech recall at $93.8\%$. Missed speech is distributed proportional to teacher and student talk time, suggesting no role-specific bias in speech retention.

\subsection{Segmentation and embedding}
Each VAD-speech region is sliced into 1.25\,s windows (0.5\,s hop), and each window is embedded with a generic pretrained speaker model (WeSpeaker \texttt{english\_resnet34}, 256-dimensional, \citet{wang2023wespeaker}).
We deliberately use \emph{one} embedder for both languages since speaker embeddings encode voice timbre rather than language, and we confirm below that the resulting features transfer.

\subsection{Clustering}
\label{sec:clustering}
Window embeddings are clustered with agglomerative clustering (cosine distance, average linkage, and distance threshold $0.5$).
We cross-validate the distance threshold jointly with the classifier in App.~\ref{app:cut} and find accuracy flat across $0.4$--$0.85$ before the clusters collapse near $0.9$. In practice, the teacher's voice is \emph{fragmented} across many clusters (often hundreds per 45-minute lesson), because content, distance, and channel drift push the same voice past the cosine threshold.

\subsection{Per-cluster structural features}
\label{sec:features}
  
For each cluster we compute a set of label-free, \emph{session-relative} features that capture the structural role of a voice without any annotation.
The most important of these is \texttt{centrality}, the cosine similarity between the cluster centroid and the duration-weighted session-global centroid. Centrality of cluster $c$ is defined as:
\begin{equation*}
  \texttt{centrality}(c) = \cos\!\big(\mu_c,\; \mu_{\text{session}}\big), \quad  \mu_c = \frac{\sum_{i\in c} d_i\, e_i}{\sum_{i\in c} d_i},
\end{equation*}
where $e_i$ is the embedding of window $i$, $d_i$ its duration, $\mu_c$ the cluster centroid, and $\mu_{\text{session}}$ the average over \emph{all} windows in the session. Because $\mu_{\text{session}}$ is duration-weighted, it is pulled toward whoever talks most, usually the teacher, so a cluster that resembles the session's dominant voice scores high.

We also compute complementary cues: talk-time share, temporal coverage and spread, turn count, cluster size, embedding consistency, and relative energy. These describe each cluster's share of the lesson's total speech duration, its temporal distribution across the lesson, and its acoustic consistency, capturing role-related patterns without requiring speaker identity. We select features by treating clusters as labeled teacher/student examples under leave-one-teacher-out cross validation, optimizing the lower per-cluster labeling accuracy across English and German. After ranking individual features, we perform greedy forward selection, adding the feature that most improves the current set. Centrality is the strongest single predictor in both languages, and additional cues provide little improvement. App.~\ref{app:features} details the feature definitions, selection procedure, and per-feature results (Figure~\ref{fig:features}).

\begin{table*}[t]
\centering
\caption{Teacher (Tch) and student (Stu) precision/recall and teacher F1 under leave-one-teacher-out cross-validation (mean~$\pm$ std).
Our pipeline outperforms enrollment-based approaches and has the lowest variance; against the external pyannote reference, it wins on German but is edged on English.
$^\dagger$pyannote (pyannote speaker-diarization-3.1, teacher chosen by our dominance rule) is a deployable CPU reference ($\sim$8\,M params) but requires license-gated diarization weights; it is scored once per corpus (no folds), with precision/recall from the aggregate teacher-vs-rest confusion.}
\label{tab:strategies}
\begin{tabular}{llccccc}
 \toprule
   & Method & TchP & TchR & TchF1 & StuP & StuR \\
  \midrule
  \StrategiesBody
  \bottomrule
\end{tabular}
\end{table*}

\subsection{Evaluation metric}
\label{sec:metric}

We compare pipeline predictions with transcript-derived role labels on $0.1$\,s frames. For each role, precision is the fraction of predicted frames that match the reference, and recall is the fraction of reference frames correctly identified. Teacher precision thus measures the purity of extracted teacher speech, while teacher recall measures how much reference teacher speech is retained within the evaluated frames. We also report F1, the harmonic mean of precision and recall. Frame-level evaluation weights speech by duration rather than by cluster count, and results are reported separately for English and German. All experiments use \emph{leave-one-teacher-out cross-validation}: in each fold, the classifier is trained on all recordings from the remaining teachers and evaluated on the held-out teacher, whose voice and session statistics are excluded from training.

\subsection{Three alternative teacher-selection strategies}
\label{sec:strategies}

We compare our pipeline against a hand-designed rule, a supervised upper bound, and an SOTA diarization tool:
\textbf{(i) Pseudo-enrollment} builds a teacher centroid from the dominant speaker in the opening minutes and labels clusters within a cosine-distance threshold as teacher. It can select the wrong voice when the teacher does not dominate the opening.
\textbf{(ii) Enrollment} uses the same threshold rule but builds its centroid from the teacher's first $30$\,s of labeled speech, providing a supervised reference that requires per-session annotation.
\textbf{(iii) pyannote speaker-diarization-3.1} is a strong external system that first diarizes an entire lesson. We pick the teacher with a dominance rule detailed in App.~\ref{app:pyannote}.


\section{Findings}
\label{sec:results}
We address six deployment questions. \textbf{Does it work?} Our pipeline outperforms enrollment approaches (Sec.~\ref{sec:res:baselines}, Table~\ref{tab:strategies}). \textbf{Why?} A session-relative feature, \texttt{centrality}, carries the signal across languages (Sec.~\ref{sec:res:mechanism}); an embedder ablation supports discourse \emph{structure} as the key cue (App.~\ref{app:embed}). \textbf{Does it generalize and deploy?} Cross-lingual transfer requires no retraining (Sec.~\ref{sec:res:transfer}), and the pipeline runs $\sim 4.8\times$ faster than a matched pyannote reference at equal memory on a commodity Android smartphone (Sec.~\ref{sec:res:deploy}).
\textbf{How does filtering affect transcription?} Filtering reduces substitutions and insertions but deletes some teacher speech, increasing teacher word error rate by $2\%$ compared to unfiltered audio (Sec.~\ref{sec:res:wer}).
\textbf{Can failures be detected?} Failures arise when teacher and student embeddings do not separate (Sec.~\ref{sec:dominance}); and they are \emph{detectable at inference}, so low-confidence sessions can be routed to human review (Sec.~\ref{sec:res:routing}). \textbf{What did not help?} Sec.~\ref{sec:res:nulls} reports three unsuccessful interventions.

\subsection{Our pipeline outperforms enrollment methods}
\label{sec:res:baselines} 

\textbf{Main findings.~~} Table~\ref{tab:strategies} compares four teacher-identification strategies under leave-one-teacher-out cross-validation across 24 English and 39 German teachers. Our Pipeline achieves teacher F1 (TchF1) of $0.81$ and $0.87$, respectively, outperforming both enrollment approaches with the lowest variability across folds. The gain comes primarily from \emph{recall}: supervised enrollment recovers just over half of teacher speech in English and two-thirds in German, whereas our pipeline reaches $0.87$ and $0.91$. Under reverberation, the teacher's voice fragments across clusters that can fall outside a fixed cosine threshold; our structural classifier evaluates each cluster and recovers more of these fragments.

\textbf{Accounting for wrong labels.~~} The human audit described in Sec.~\ref{sec:data} reveals substantial transcript errors. Using human judgments and weighting each sampling stratum by its corpus duration raises estimated teacher F1 from $0.84$ to $0.94$ (App.~\ref{app:gold}). The correction is much larger for precision than recall: nearly two-thirds of the speech flagged as teacher false positives is actually teacher speech. Errors also persist where the pipeline and transcript agree. Teacher-labeled agreement regions are largely reliable, but $56.5\%$ of the audited student-labeled agreement regions contain teacher speech or non-speech. The student channel's apparent weakness in Table~\ref{tab:strategies} therefore reflects substantial reference-label noise. Since the audit covers full transcript intervals, its correction is an upper bound for the onset-restricted main evaluation.

\textbf{Pyannote comparison.~~} Pyannote reaches TchF1 of $0.84$ in English and $0.82$ in German, slightly outperforming our pipeline on English but trailing it on German. Both systems use the same WeSpeaker ResNet34 embedder. What distinguishes our pipeline is not model size, but that it recovers the teacher from a \emph{generic} embedder and session-relative structure, without gated diarization weights or the neural segmentation stage, pyannote runs ahead of clustering.

Overall, our approach generalizes to unseen teachers and improves recall over supervised enrollment through structural cluster features. App.~\ref{app:embed} tests whether this advantage persists with a stronger embedder.

\subsection{Why it works: \texttt{centrality} is the dominant, transferable cue}
\label{sec:res:mechanism}

\begin{figure*}[t]
\centering
\includegraphics[width=\linewidth]{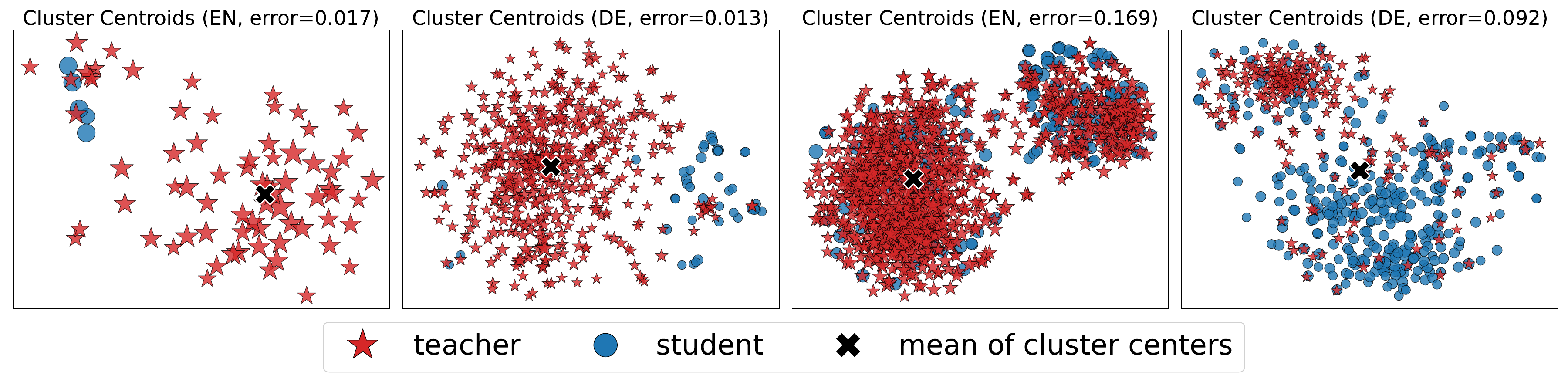}
\vspace*{-.75cm}
\caption{t-SNE embeddings of \emph{cluster centroids}, coloured by majority ground-truth role. \textbf{Two good cases} (left two): the teacher clusters are tight, and the global mean falls inside the teacher mass, so a single global centroid recovers the teacher. \textbf{Two bad cases} (right two): the teacher clusters are fragmented, and the global mean drifts off the teacher mass, so the same proxy fails.}
\label{fig:tsne}
\end{figure*}

Figure~\ref{fig:tsne} visualizes cluster centroids coloured by their majority ground-truth role. In successful recordings, teacher clusters are compact and well separated from student clusters, placing the session-global mean near the teacher clusters. Consequently, \texttt{centrality} assigns high scores (cosine similarity) to multiple teacher fragments. In difficult recordings, teacher and student clusters overlap, weakening this cue.

Single-feature ablations support this explanation: \texttt{centrality} alone achieves teacher F1 of $0.85$, and adding other features through greedy forward selection provides little improvement (App.~\ref{app:features}, Figure~\ref{fig:features}). Embedding using $6\times$ larger ResNet221 also improves F1 by less than $0.01$ in either language (App.~\ref{app:embed}). These support using a compact embedder with session-relative features for on-device inference. 

\subsection{Cross-lingual transfer is free}
\label{sec:res:transfer}

The main experiments train a single role classifier jointly on German and English recordings. Training on either language alone and evaluating on a held-out teacher (same language) and lessons in the other language yields the same English/German TchF1 of $0.81/0.87$ (Table~\ref{tab:strategies}). Thus, the session-relative cue captured by \texttt{centrality} transfers across these corpora despite different recording conditions and classroom practices, allowing deployment in the other language without modification or additional annotation.

\subsection{Deployability: runtime, footprint, and clustering robustness}
\label{sec:res:deploy}

The full pipeline (VAD, embedding, clustering, and role classification) runs locally on CPU, allowing audio and intermediate speaker representations to remain on the device. Having established competitive accuracy against pyannote (Sec.~\ref{sec:res:baselines}), we compare their runtime and memory requirements.

\begin{figure}[t]
\centering
\includegraphics[width=\linewidth]{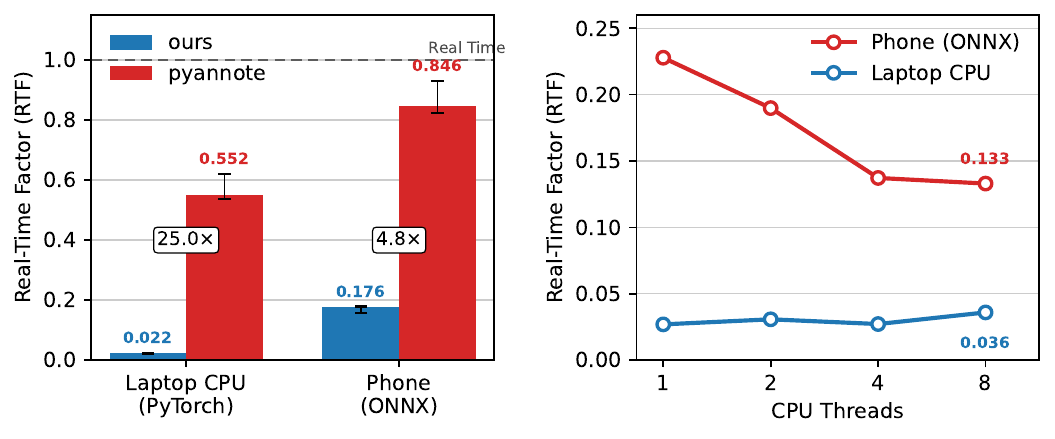}
\caption{\textbf{Efficiency of the proposed pipeline versus \texttt{pyannote}.}
\textbf{Left:} Real-time factor at matched accuracy (recording~A, single thread); error
bars span the recorded min--max, the dashed line is real time, the boxed figure is the
within-environment speed-up. \textbf{Right:} Thread scaling (recording~B): the phone nearly
halves its RTF from one to eight threads as embedding parallelises, while the laptop is
already fast enough that RTF is flat. Lower is faster.}
\label{fig:bench}
\end{figure}

We benchmark both pipelines end-to-end on a laptop CPU and a commodity Android smartphone using matched inference settings. Fig.~\ref{fig:bench} shows the results, and App.~\ref{app:bench} provides the hardware and protocol. On the laptop, our pipeline achieves a real-time factor (RTF) of $0.022$: approximately $25\times$ faster than pyannote and $45\times$ faster than real time. On the phone, where the baseline is itself an optimised ONNX build(\citep{sherpaonnx}), it is $4.8\times$ faster. Both approaches use the same RSS peak resident memory (872 MB on laptop vs 356 MB on phone (Figure~\ref{fig:bench})). We also benchmark the parallelisability of our pipeline by increasing the number of threads (Table~\ref{tab:scaling}, Fig.~\ref{fig:bench}). We find that while there are efficiency gains on the phone with RTF nearly halving on 8 threads, the efficiency on laptop CPU stays flat or decreases.


\subsection{Downstream transcription analysis}
\label{sec:res:wer}

We assess how role filtering affects teacher transcription by running Whisper-large-v2 \citep{radford2023robust} on all 292 recordings in fixed $30$\,s chunks. We compare unfiltered audio with audio retaining only spans the pipeline labels as teacher. Both conditions are evaluated against the teacher's transcript turns using the teacher-word scoring protocol in App.~\ref{app:wer}.

Filtering modestly increases average teacher word error rate (WER), from $0.26$ to $0.28$ across the two languages. The effect differs by corpus: English WER rises from $0.160$ to $0.213$, with most recordings worsening, while German performance remains essentially unchanged.

Filtering reduces substitutions and insertions in both corpora, but also removes some genuine teacher speech. The resulting deletions outweigh the recognition gains, particularly in English. Frequently deleted words include fillers, backchannels, and discourse markers, e.g., \emph{okay}, \emph{right}, \emph{ja}, and \emph{mhm} (Figure~\ref{fig:delwords}). Although they carry little content, their loss may still matter for analyses of classroom interaction.


\begin{table}[t]
\centering
\caption{Embedding-level oracle vs.\ our pipeline. Oracle is a supervised logistic probe on the raw embeddings (the ceiling extractable from the representation), and GT sil.\ is the cosine silhouette of the \emph{true} teacher/student partition. Overall, the two match; only the low-accuracy tail leaves a recoverable signal, and it remains a low-separability regime.}
\label{tab:oracle}
\small
\begin{tabular}{lrcccc}
\toprule
F1 stratum & $n$ & pipeline & oracle & $\Delta$ & GT sil. \\
\midrule
$<0.70$        & $12$  & $0.63$ & $0.78$ & $+0.15$ & $0.09$ \\
$0.70$--$0.80$ & $34$  & $0.76$ & $0.78$ & $+0.02$ & $0.05$ \\
$0.80$--$0.90$ & $116$ & $0.86$ & $0.85$ & $-0.01$ & $0.11$ \\
$\ge 0.90$     & $128$ & $0.94$ & $0.92$ & $-0.02$ & $0.18$ \\
\midrule
all            & $290$ & $0.87$ & $0.87$ & $-0.00$ & $0.13$ \\
\bottomrule
\end{tabular}
\end{table}

\subsection{When the pipeline fails: separability sets the ceiling}
\label{sec:dominance}

Per-recording accuracy depends strongly on how clearly teacher and student speech separate in the embedding space. We measure this without reference labels using the \emph{silhouette} of the predicted role partition: for each embedding, the difference between its mean cosine distance to the other role and to its own role, normalized by the larger of these distances and averaged across embeddings. Teacher F1 correlates strongly with this measure ($\rho=0.76$) and drops sharply below a silhouette of $0.10$ (Figure~\ref{fig:dominance}), linking the poorest-performing recordings to weak separation between the two roles.

To examine how much signal remains recoverable, we train a supervised, within-recording linear probe on the raw $256$-dimensional window embeddings. Using ground-truth role labels, this probe assesses how well a linear classifier can distinguish teacher from student speech directly from the embeddings (Table~\ref{tab:oracle}). Its overall teacher F1 matches the pipeline at $0.87$. In the lowest-performing group, however, it improves F1 from from $0.63$ to $0.78$, indicating that our pipeline does not recover some linearly separable signal. Yet, even there, it reaches only $0.78$, i.e., some recordings are hard because the true embeddings barely separate.

\subsection{Failures are detectable at inference: routing to human review}
\label{sec:res:routing}

The difficult recordings in Sec.~\ref{sec:dominance} can often be flagged without reference labels. Two inference-time diagnostics correlate strongly with teacher F1: the \emph{classifier margin}, the mean distance of the teacher probability from $0.5$, and the \emph{predicted teacher share}, the fraction of speech assigned to the teacher (Spearman $\rho = 0.82$ and $0.73$; see App.~\ref{app:diag}, Table~\ref{tab:diag}).

Classifier margin is particularly useful for review routing. Flagging the lowest-margin $20\%$ of recordings captures $85\%$ of those with teacher F1 below $0.70$ and raises mean F1 among retained recordings from $0.87$ to $0.92$. Predicted teacher share shows a similar association: recordings with lower shares tend to have lower accuracy (Figure~\ref{fig:predshare}). These diagnostics offer a practical way to prioritize human review; App.~\ref{app:diag} compares their predictive strength.

\subsection{What does not help}
\label{sec:res:nulls}
Three interventions do not change the results significantly. We defer the details of these negative results to appendices.

\textbf{Accumulating a teacher voiceprint across lessons.} ~~ 
Pooling predicted teacher clusters from earlier lessons provides little benefit: teacher F1 increases by only $0.01$ in German and remains unchanged in English. Similarity to the accumulated voiceprint closely tracks similarity to the session-global mean, suggesting that it adds little information beyond \texttt{centrality}. Label-free accumulation also matches supervised enrollment, supporting its use when a cross-lesson voiceprint is available to avoid enrollment (App.~\ref{app:online}, Figure~\ref{fig:online}).

\textbf{Masking mixed windows.} ~~ Roughly $9\%$ of English and $4\%$ of German speech are flagged by pyannote/segmentation-3.0 as overlapping speech. Excluding these windows from the feature calculations changes teacher F1 by at most $0.002$ across separability strata. This suggests that the aggregate features are insensitive to this masking intervention, so we retain the simpler pipeline without overlap detection (App.~\ref{app:mask}).

\textbf{Varying the onset-trust window.~~}
As noted in Sec.~\ref{sec:data}, transcripts mark turn onsets rather than true speech offsets. So, we restrict evaluation to a short window after each onset to reduce incorrect attribution of later pauses, backchannels, and other-speaker speech. Re-scoring the predictions with onset-trust windows of $\{2,4,8\}$\,s and full transcript intervals preserves the strategy ranking. Student precision falls as the window widens, consistent with longer interval tails introducing incorrect role labels. The main comparison is therefore robust to the window choice, while student precision remains sensitive to transcript span inflation (App.~\ref{app:headsweep}).

\section{Related work}
\textbf{Speaker diarization.} ~~ 
Speaker diarization determines \emph{who spoke when} \citep{anguera2012review, park2022review}, commonly through speaker embeddings and clustering \citep{snyder2018xvectors, desplanques2020ecapa, wang2023wespeaker, landini2022vbx} or end-to-end neural approaches \citep{fujita2019eend, horiguchi2020eda, plaquet2023powerset, park2024sortformer}. Related systems use enrollment audio to detect target speakers \citep{medennikov2020tsvad} or combine diarization with ASR \citep{bain2023whisperx}. Our task adds role assignment to this setting: the pipeline must identify teacher speech without enrollment and operate within a smartphone CPU budget.

\textbf{Classroom and child-adult audio.} ~~ 
Teacher-talk analysis spans wearable adult/child tagging \citep{xu2009lena, ganek2018lena} and classroom-discourse analytics that rely on teacher/student segmentation \citep{owens2017dart, wang2014instructional, donnelly2017words, kelly2018authenticity, schlotterbeck2021tarta}. Short turns, cross-talk, and limited representation of children's speech in pretrained systems make this setting challenging \citep{kothalkar2024childadult}. Related work explores simulated child-adult conversations to reduce dependence on sensitive recordings \citep{xu2025simulated}, therapist-role identification in psychotherapy \citep{flemotomos2020linguistically}, and video-assisted classroom speaker identification \citep{chen2019multimodal, chrzan2026multimodal}. Our pipeline uses audio alone and assigns roles through session-relative features.

\textbf{Closest systems.} ~~ 
\citet{wang2025jedm} study a cluster-ing-based pipeline for classroom group discussions using ECAPA-TDNN embeddings, Whisper-derived segmentation, and enrollment audio. Our pipeline instead combines dedicated VAD with explicit teacher-role classification and requires no enrollment. \citet{khan2025multi} address classroom noise through a multi-stage pipeline combining denoising, neural VAD, and ASR timestamps, resulting in a server-side and GPU-scale setting. We focus on a compact inference path that runs faster than real time on a commodity smartphone CPU. Speech-privacy research also studies federated learning \citep{granqvist2020federated, woubie2021federated} and anonymisation \citep{sumer2020anonymisation}. Our deployment approach complements these directions by keeping audio and speaker representations on the device during inference.

\section{Discussion and Conclusion}
\label{sec:discussion}

Our results suggest that teacher-role recovery can rely on session-relative structure without speaker enrollment. Centrality supplies most of the predictive signal, transfers between the two evaluated languages, and gains little from a larger embedder or accumulated teacher voiceprints. This supports a compact, on-device design for classroom role segmentation, although its value for educational feedback remains to be established in operational use.

Reliability remains conditional on teacher/student acoustic separability. Low classifier margins identify many difficult sessions, supporting selective human review rather than uniformly trusting every output. Filtering also introduces a downstream trade-off: average teacher WER rises from $0.26$ to $0.28$, with greater degradation in English. Deleted backchannels and discourse markers may matter for interaction analysis even when they contribute little lexical content.

Our findings are restricted to two corpora in which language and microphone regime vary together. The timestamp audit also shows that transcript-derived labels are imperfect to measure accuracy. Future work should prioritize precisely timed role annotations, prospective evaluation of thresholds, and testing in more diverse classrooms. Overall, the pipeline provides a practical basis for local teacher/student segmentation, with explicit reliability checks and task-specific validation remaining essential for deployment.

\section*{Ethical Statement}

We use two existing research corpora, collected under their originating studies' informed-consent procedures and ethics approvals and accessed under their data-use agreements. These recordings involve minors and are not redistributed. Our analysis distinguishes discourse roles without identifying individual students. Segmentation runs locally, allowing raw recordings and intermediate speaker embeddings to remain on the device. However, filtering does not anonymize speech: teacher segments remain identifiable, and classification errors can retain student speech. 

The pipeline is intended to support aggregate classroom-discourse analysis for research and formative feedback. Its outputs should not serve as standalone measures of teaching quality or support high-stakes decisions about teachers or students. Errors may distort talk-time estimates and remove meaningful speech, while performance beyond the evaluated classrooms remains uncertain. Deployments should limit data access and retention and provide human oversight. Our inference-time diagnostics can prioritize potentially unreliable sessions for review (Sec.~\ref{sec:res:routing}), but do not guarantee accurate role assignments or prevent privacy-sensitive errors.

\paragraph{Acknowledgements}
Çağatay Yıldız is a member of the Machine Learning Cluster of Excellence, funded by the Deutsche Forschungsgemeinschaft (DFG, German Research Foundation) under Germany’s Excellence Strategy – EXC number 2064/1 – Project number 390727645. This research utilized compute resources at the Tübingen Machine Learning Cloud, DFG FKZ INST 37/1057-1 FUGG.

\bibliography{main}   

\clearpage
\appendix


\begin{table*}[h]
\centering
\caption{Overview of appendix sections and supplementary analyses.}
\label{tab:appendix-overview}
\begin{tabular}{@{}p{0.15\textwidth}p{0.57\textwidth}p{0.22\textwidth}@{}}
\toprule
\textbf{Item} & \textbf{Description} & \textbf{Complements} \\
\midrule
Figure~\ref{fig:dominance} & Relationship between teacher/student embedding separability and per-recording teacher F1 & Sec.~\ref{sec:dominance} \\
\midrule
App.~\ref{app:pyannote} & Details of how we map pyannote annotations to teacher/student labels & Secs.~\ref{sec:strategies},\ref{sec:res:baselines} \\
App.~\ref{app:gold} & Human audit of transcript labels: sampling, annotation, and duration-weighted corrected metrics & Secs.~\ref{sec:data}, \ref{sec:res:baselines} \\
App.~\ref{app:headsweep} & Sensitivity of teacher F1 and student precision to the onset-trust window & Secs.~\ref{sec:data}, \ref{sec:res:nulls} \\
App.~\ref{app:embed} & ResNet34 versus ResNet221 embeddings for our pipeline and supervised enrollment & Sec.~\ref{sec:res:mechanism} \\
App.~\ref{app:cut} & Accuracy and cluster-count sensitivity to the clustering distance threshold & Sec.~\ref{sec:clustering} \\
App.~\ref{app:features} & Structural feature definitions, feature selection, and single-feature performance & Secs.~\ref{sec:features}, \ref{sec:res:mechanism} \\
\midrule
App.~\ref{app:diag} & Label-free session diagnostics and their association with teacher F1 & Sec.~\ref{sec:res:routing} \\
App.~\ref{app:online} & Cross-lesson teacher accumulation and its limited benefit beyond centrality & Sec.~\ref{sec:res:nulls} \\
App.~\ref{app:mask} & Effect of excluding overlap-detected windows from feature computation & Sec.~\ref{sec:res:nulls} \\
\midrule
App.~\ref{app:bench} & Runtime implementations, benchmark protocol, hardware, and thread scaling & Sec.~\ref{sec:res:deploy} \\
App.~\ref{app:wer} & Teacher transcription protocol, corpus-level error breakdowns, and frequently deleted words & Sec.~\ref{sec:res:wer} \\
\bottomrule
\end{tabular}
\end{table*}

\clearpage

\paragraph{A note on data preprocessing.} We start by noting the audio preprocessing. All audio is decoded to 16\,kHz mono \texttt{float32}. Video sources are converted with a fixed \texttt{ffmpeg} recipe (\texttt{-vn -ac 1 -ar 16000 -c:a pcm\_s16le}). For the English close-mic case, we additionally check for a separate loud channel that would give teacher activity directly; our recordings are mono, so the mixed-down path is used throughout. 

\section{Assigning roles to \texttt{pyannote} speakers.}
\label{app:pyannote}
\texttt{pyannote} partitions a lesson into anonymous speakers but does not identify their roles. We therefore map its output to teacher and student labels using a label-free \emph{dominance rule}. Each VAD window is assigned to the speaker with the greatest temporal overlap. We then score each speaker using $D = \texttt{talk\_time\_frac} \times \texttt{coverage} \times turn\_count$. For English recordings, where teachers wear a close-talk microphone, we additionally multiply this score by a shifted \texttt{energy\_z} term. We decided this rule after cross validating the individual factors and combining pairs/triplets.

The speaker with the highest score is labeled teacher, and all remaining speakers are labeled student. The product favors speakers who combine three properties: they account for a large share of speech, participate throughout the lesson, and take many turns. Together, these cues help distinguish the teacher from a talkative student who may score highly on one component but is less likely to score highly on all three. This rule is particularly suitable for \texttt{pyannote}, whose output aims to group each speaker's speech under a single identity, making the selection of one highest-scoring speaker a natural way to assign the teacher role.

\begin{figure}[t]
\centering
\includegraphics[width=.99\linewidth]{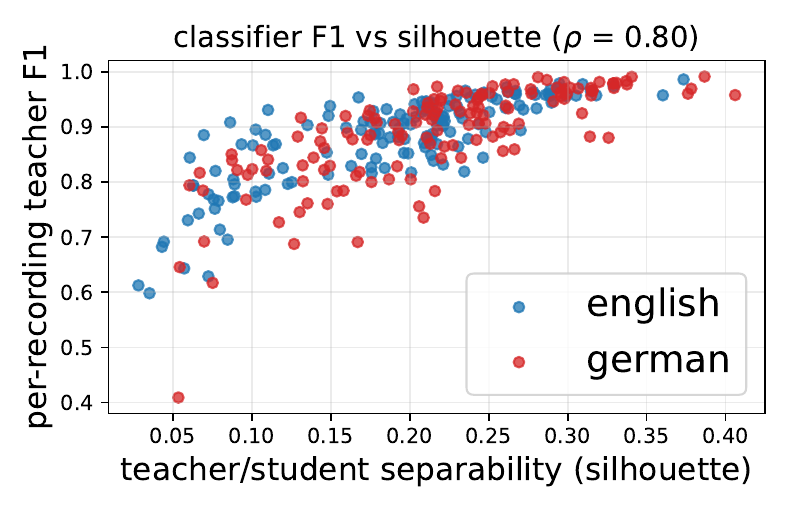}
\vspace{-.25cm}
\caption{Per-recording tchF1 against teacher/student embedding separability (silhouette). The relationship is tight (Spearman $\rho = 0.80$) and tchF1 collapses once separability falls below $\approx 0.10$: the low-F1 recordings are the ones whose teacher and student voices the embeddings fail to pull apart.}
\label{fig:dominance}
\end{figure}

\section{A human-labeled audit of the ground truth}
\label{app:gold}

Sections~\ref{sec:data} and \ref{sec:res:baselines} claim that a large share of the residual error in Table~\ref{tab:strategies} is label error. This appendix gives the measurement behind that claim.

\paragraph{Design.}
We sampled $5{,}000$ clips from $202$ of our recordings ($134$ German, $68$ English) and had a single trained annotator label each one by ear as teacher/adult (\texttt{t}), student/child (\texttt{s}), non-speech (\texttt{n}), or unclear (\texttt{u}), with a separate binary flag for audible overlap. The annotator saw neither the model output nor the transcript, and clip order was shuffled, so no stratum was identifiable during labeling. $4{,}998$ clips were returned labeled.

Sampling is stratified into six strata, chosen so that the estimator can correct the \emph{agreement} regions rather than only the disagreements. Two strata cover the frames where the pipeline and the transcript disagree (\textbf{FP}: pipeline teacher, transcript student; \textbf{FN}: pipeline student, transcript teacher); two are agreement controls (\textbf{ctrlT}, \textbf{ctrlS}), which measure the transcript's error rate where nothing flags a problem; and two cover the regions the VAD discards as silence, split at the median frame energy (\textbf{silHigh}, \textbf{silLow}) so that quiet regions are measured rather than assumed non-speech. Sampling is budgeted by clip \emph{count}, since annotation effort scales with the
number of decisions rather than with audio duration.

\begin{table*}[t]
\centering
\caption{The six sampling strata: total duration in the corpus, clips sampled, and the duration-weighted distribution of human labels within each stratum. \textbf{ov} is the fraction of \emph{speech-labeled} clip time carrying the overlap flag. Rows sum to $1$ up to the unused
\texttt{u} label (see the caveat below).}
\label{tab:gold}
\small
\setlength{\tabcolsep}{4pt}
\begin{tabular}{llrrrrrr}
\toprule
stratum & pipeline / transcript & tot (ks) & clips & t & s & n & ov \\
\midrule
FP      & teacher / student  & $26.5$  & $950$  & $0.63$ & $0.11$ & $0.26$ & $0.12$ \\
FN      & student / teacher  & $35.0$  & $1520$ & $0.24$ & $0.16$ & $0.60$ & $0.15$ \\
ctrlT   & teacher / teacher  & $165.4$ & $559$  & $0.96$ & $0.00$ & $0.03$ & $0.02$ \\
ctrlS   & student / student  & $46.4$  & $560$  & $0.18$ & $0.44$ & $0.38$ & $0.16$ \\
\midrule
silHigh & VAD silence, loud  & $58.2$  & $870$  & $0.09$ & $0.05$ & $0.86$ & $0.06$ \\
silLow  & VAD silence, quiet & $126.4$ & $539$  & $0.07$ & $0.02$ & $0.91$ & $0.10$ \\
\bottomrule
\end{tabular}
\end{table*}

\paragraph{Estimator.}
Each stratum is a uniform sample of a known total duration, so a duration-weighted fraction within
a stratum extrapolates to the corpus by scaling with that total. Writing $T(\cdot)$ for a stratum's
total seconds and $f_x(\cdot)$ for the fraction of its sampled clip-time labeled $x$, the corrected
teacher confusion is
\begin{align*}
\mathrm{TP} &= T(\text{ctrlT})f_t(\text{ctrlT}) + T(\text{FP})f_t(\text{FP}),\\
\mathrm{FP} &= T(\text{ctrlT})f_s(\text{ctrlT}) + T(\text{FP})f_s(\text{FP}),\\
\mathrm{FN} &= T(\text{ctrlS})f_t(\text{ctrlS}) + T(\text{FN})f_t(\text{FN}),
\end{align*}
and symmetrically for the student role. VAD recall compares the retained speech total against the
speech recovered from the two silence strata,
$T(\text{silHigh})f_{\mathrm{sp}}(\text{silHigh}) + T(\text{silLow})f_{\mathrm{sp}}(\text{silLow})$
with $f_{\mathrm{sp}} = f_t + f_s$. Confidence intervals are the $2.5$/$97.5$ percentiles of $1000$
bootstrap resamples drawn within each stratum.

\begin{table}[t]
\centering
\caption{Human-corrected metrics. ``transcript'' scores the same recordings with the transcript taken as truth; ``human'' treats the annotator as truth. Both are pooled across corpora and scored over full transcript intervals, so the transcript column is the like-for-like reference.}
\label{tab:goldres}
\small
\begin{tabular}{lccl}
\toprule
 & transcript & human & $95\%$ CI \\
\midrule
teacher precision & $0.86$ & $0.98$ & $[0.97,0.99]$ \\
teacher recall    & $0.83$ & $0.91$ & $[0.90,0.92]$ \\
teacher F1        & $0.84$ & $\mathbf{0.94}$ & $[0.94,0.95]$ \\
student F1        & $0.60$ & $0.71$ & --- \\
\midrule
transcript error, ctrlT & --- & $0.035$ & $[0.023,0.048]$ \\
transcript error, ctrlS & --- & $0.565$ & $[0.513,0.621]$ \\
VAD recall              & --- & $0.928$ & $[0.910,0.945]$ \\
\bottomrule
\end{tabular}
\end{table}

\paragraph{Results.}
Table~\ref{tab:goldres} collects the estimates. Correcting the labels moves teacher F1 from $0.84$
to $0.94$, i.e.\ roughly two thirds of the counted teacher error is not model error, and the
correction lands on precision rather than recall. The two control strata explain why: where the
transcript and the model agree on \texttt{teacher} it is right $96.5\%$ of the time, but where they
agree on \texttt{student} the label is wrong on $56.5\%$ of the audited time---only $44\%$ of it is
a student speaking, the rest being teacher speech ($18\%$) or non-speech ($38\%$). The
\texttt{student} label is thus closer to a residual class than to a positive one, which depresses
every student-side number in the main text.

The silence strata bound what the VAD discards: $92.8\%$ of speech is retained, and the missed
$7\%$ divides between the roles in proportion to their talk time ($72.8\%$ of it teacher, against a
$73.3\%$ teacher base rate), so it does not bias the teacher/student comparison. 

\paragraph{Caveats.}
Two limitations apply to our inspection. First, the labels come from a \emph{single} annotator. We report no inter-annotator agreement; the direction of the result replicates a smaller pilot audit by one of the authors on disjoint clips, but the magnitudes should be read as one rater's judgment. Second, the pool was drawn against ground truth over full transcript intervals, whereas the main results trust only the first $h=4$\,s after each onset (Section~\ref{sec:metric}). Since that window already removes part of the same label noise, the correction reported here is an upper bound on what applies to Table~\ref{tab:strategies}.

\section{Sensitivity to the ground-truth onset-trust window}
\label{app:headsweep}

The evaluation metric (Section~\ref{sec:metric}) trusts only the first $h=4$\,s of speech after each transcript onset as ground truth since the transcripts are onset-anchored. More concretely, an interval marks \emph{when} a turn begins, not how long the speaker talks, so the interval tail silently contains pauses, backchannels, and overlapping speeches from the other party. The $4$\,s window is therefore a hyperparameter, and a natural worry is that the reported results are an artifact of that particular choice. To rule this out, we re-score the \emph{same} per-fold predictions against ground truth computed at $h\in\{2,4,8\}$\,s and at the full interval ($h\to\infty$, i.e.\ every frame from one onset to the next is attributed to the named speaker). Figure~\ref{fig:headsweep} reports teacher F1 and student precision for all three strategies, leading to two clear patterns:

\begin{figure*}[t]
\centering
\includegraphics[width=0.995\linewidth]{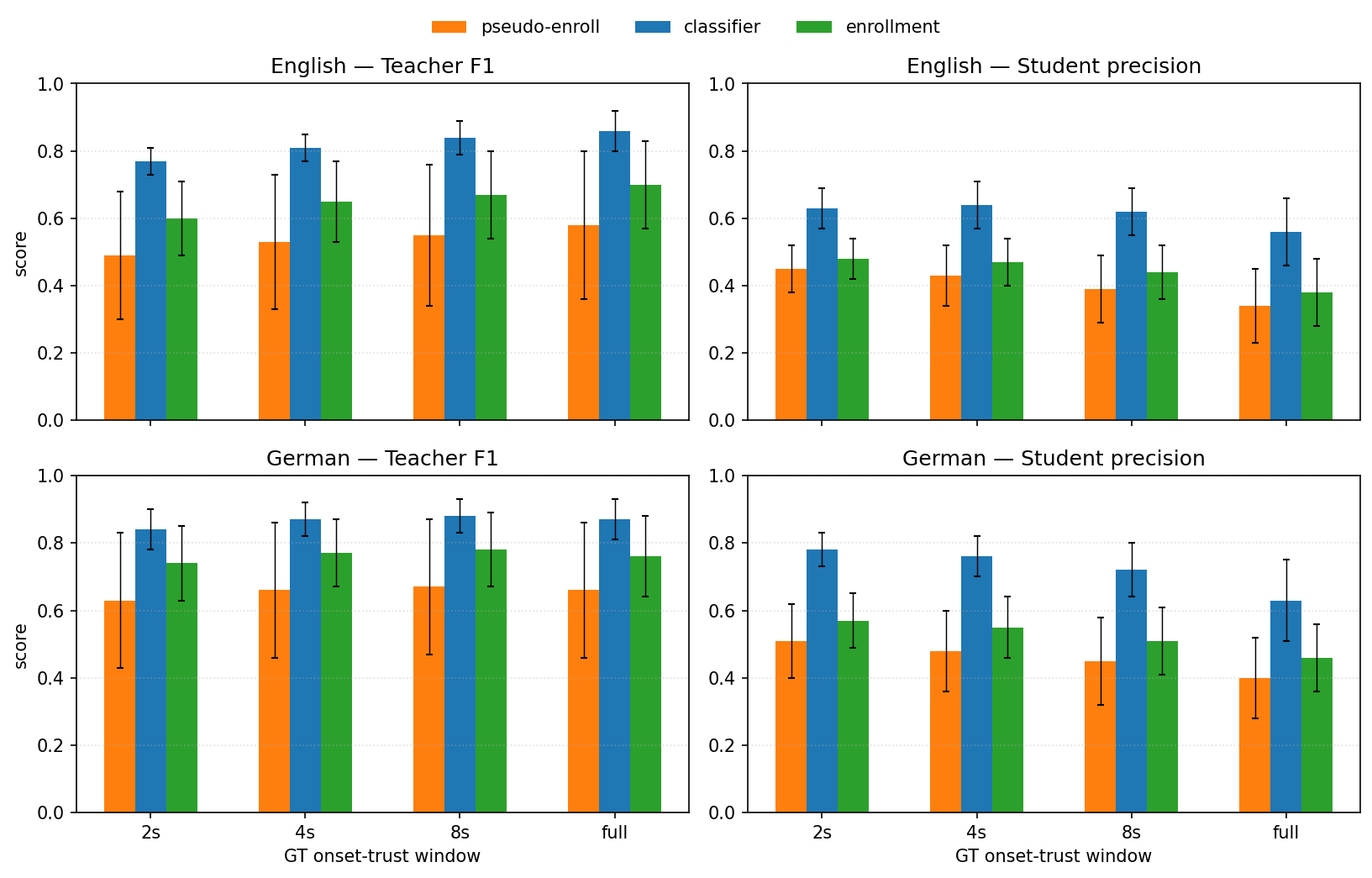}
\caption{Effect of the ground-truth onset-trust window on the evaluation. Rows are languages, columns are teacher F1 and student precision; within each panel the bars are grouped by window ($2$/$4$/$8$\,s and full interval) and coloured by strategy (error bars: std across leave-one-teacher-out folds). \textbf{Teacher F1 (left) is flat-to-rising} as the window widens, and the ranking of strategies is preserved at every window. \textbf{Student precision (right) falls monotonically}. The signature of inflated interval tails being folded into the dominant speaker's ground truth.}
\label{fig:headsweep}
\end{figure*}

\textbf{Teacher F1 is essentially flat.~~} For our pipeline, it moves from $0.77$ to $0.86$ (English) and $0.84$ to $0.87$ (German) as $h$ grows from $2$\,s to the full interval. More importantly, the ranking of the three strategies is identical at every window: our pipeline wins, enrollment is the next best, and then pseudo-enrollment. The main-text conclusions do not depend on the value of $h$.

\textbf{Student precision, by contrast, falls monotonically~~.} As the window widens, our pipeline drops from $0.63$ to $0.56$ (English) and $0.78$ to $0.63$ (German) between $h=2$\,s and the full interval. This is exactly the inflated-tail effect the metric is designed to avoid. As $h$ grows, more of each interval's tail, which in reality contains silence and short turns by the other speaker, is credited to the named speaker in the ground truth. Frames the pipeline \emph{correctly} assigns to the student are then scored against a label that has been overwritten as teacher, so student precision is penalized for being right. Full-interval scoring, therefore, does not measure the system more realistically; it measures the system against noisier labels, under-crediting student purity. The $h=4$\,s window sits where teacher recovery has already saturated, but tail contamination is still small, and we adopt it for the main results on that basis. We note that this robustness check uses the existing onset-anchored transcripts; a frame-accurate evaluation of sustained turns would require a small subset re-labeled with true speech offsets, which we leave to future work.

\section{Embedding-model robustness: ResNet34 vs.\ ResNet221}
\label{app:embed}

All results in the main text use a compact VoxCeleb ResNet34 speaker embedder, chosen for its small footprint. A natural question is whether the method is bottlenecked by embedding capacity. In particular, whether the supervised enrollment ceiling and the gap between it and our label-free pipeline would shrink with a stronger model. To probe this, we rebuild both caches with a $6\times$ larger embedder from the same family, VoxCeleb ResNet221, and re-run the entire leave-one-teacher-out evaluation. 

The main finding is that \textbf{our pipeline is essentially invariant to the embedder} (Table~\ref{tab:embed-clf}): teacher F1 changes by less than $0.01$ in either language, and every other cell moves within fold-level noise. By contrast, \textbf{every cosine-distance strategy improves} with the larger model. The clearest gain is for supervised enrollment (Table~\ref{tab:embed-enroll}), whose teacher recall rises by $+0.05$ (English) and $+0.04$ (German) and F1 by $+0.03$/$+0.02$; pseudo-enrollment shifts in the same direction. The better embedding flows into exactly the mechanisms that decide membership by voiceprint geometry, and not into the structural-feature classifier.

\begin{table*}[t]
\centering
\caption{Embedding-model ablation for our pipeline: ResNet34 vs.\ a $6\times$ larger ResNet221, frame-level mean~$\pm$ std across leave-one-teacher-out folds ($h=4$\,s). The classifier is invariant to the embedder.}
\label{tab:embed-clf}
\begin{tabular}{llccccc}
\toprule
 & Embedder & TchP & TchR & TchF1 & StuP & StuR \\
\midrule
\EmbedClfBody
\bottomrule
\end{tabular}
\end{table*}

\begin{table*}[t]
\centering
\caption{Same ablation for the supervised \emph{enrollment} upper bound, which decides cluster membership by cosine distance to a confirmed-teacher centroid. Unlike our pipeline, it benefits from the stronger embedder (especially in recall) yet still trails our pipeline of Table~\ref{tab:embed-clf} by $\sim$$0.10$ joint F1.}
\label{tab:embed-enroll}
\begin{tabular}{llccccc}
\toprule
 & Embedder & TchP & TchR & TchF1 & StuP & StuR \\
\midrule
\EmbedEnrollBody
\bottomrule
\end{tabular}
\end{table*}

Two conclusions follow these observations. First, the larger embedder \emph{raises} the supervised geometry ceiling, and the label-free classifier still beats it by roughly $0.10$ F1 and $0.22$ recall. Thus, we conclude that the recall gap reported in Section~\ref{sec:results} is not an artefact of a weak embedder, but a property of voiceprint-distance aggregation itself: a stronger model cannot close it. Second, that the structural classifier is unmoved while clustering thresholds were left un-recalibrated indicates its inputs are robust to the embedding change, consistent with the features being session-relative rather than tied to any particular embedding geometry. Both observations reinforce the central claim that the teacher is recoverable from discourse structure rather than acoustic detail. In practice, they justify using the smaller ResNet34, which matches the larger model on the deployment path at a fraction of the cost.

\section{Sensitivity to the clustering cut}
\label{app:cut}

The agglomerative distance threshold (Section~\ref{sec:clustering}) is the most important clustering parameter. Setting it too low shatters each speaker into noise-level micro-clusters, and too high merges teacher and student into one cluster that no role rule can separate. We sweep the cut over $\{0.4,0.5,0.6,0.7,0.75,0.8,0.85,0.9\}$. At each cut, we re-cluster, regenerate the structural features and per-cluster labels, and run the full leave-one-teacher-out evaluation. We score each language by the symmetric \emph{macro role-F1} (the mean of teacher and student F1) and take the best decision threshold $\tau$ per cut.

\begin{figure}[t]
\centering
\includegraphics[width=0.995\linewidth]{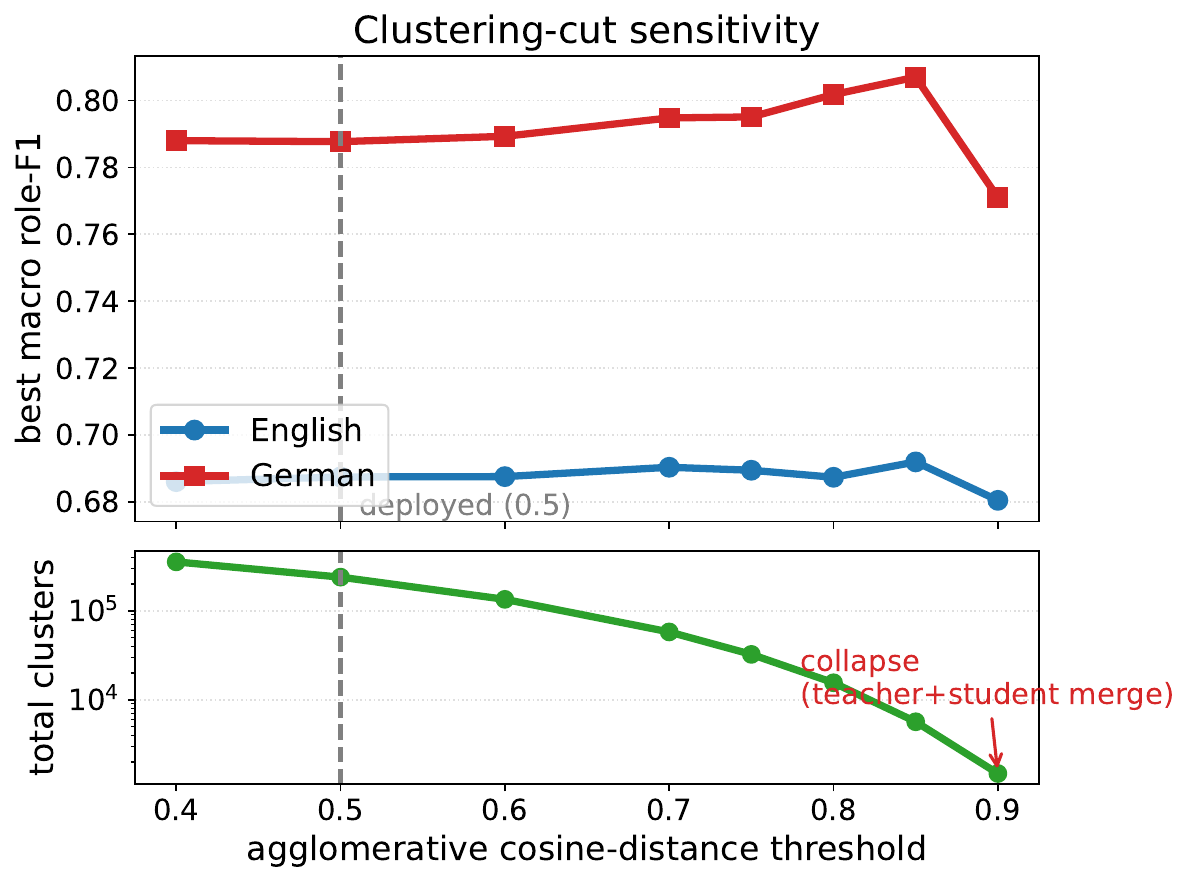}
\caption{Clustering-cut sensitivity. \textbf{Top:} best macro role-F1 (mean of teacher and student F1, leave-one-teacher-out) per language as the cosine-distance threshold varies. Accuracy is flat to within $\sim$$0.006$ across the entire $0.4$--$0.85$ range for both languages. \textbf{Bottom:} the total cluster count (log scale) falls from $3.5{\times}10^5$ to $5.7{\times}10^3$ over the same range. At $0.9$ the clusters collapse as teacher and student voices merge, and macro F1 drops sharply. The deployed $0.5$ (dashed) sits in the middle of the flat plateau, far from the cliff.}
\label{fig:cutsweep}
\end{figure}

Figure~\ref{fig:cutsweep} shows the result. The macro objective is \textbf{flat to within $\sim$$0.006$ across the whole $0.4$--$0.85$ range}. English moves only between $0.686$ and $0.692$, German between $0.78$ and $0.81$ despite the cluster count falling by more than two orders of magnitude (from $\sim$$3.5{\times}10^5$ clusters at $0.4$ to $\sim$$5.7{\times}10^3$ at $0.85$). This is the same robustness shown by the embedder ablation from a different angle: the session-relative structural features the classifier relies on survive substantial changes in clustering granularity. Only at $0.9$, the recordings collapse to $\sim$$5$ clusters per recording, making the teacher and student no longer separable. These observations confirm that the cut matters as a \emph{safety floor} rather than an accuracy lever. We therefore keep the calibrated $0.5$, which sits in the middle of the plateau and well clear of the collapse, and decline the $\le0.005$ macro-F1 that the coarsest safe cuts would buy at the cost of operating one step from breakdown.

\section{Feature set, selection, and per-feature analysis}
\label{app:features}

This appendix gives the full structural feature set that complements \texttt{centrality}
(Section~\ref{sec:features}), the selection procedure summarised in Section~\ref{sec:features},
and the per-feature results that establish centrality as the dominant, language-transferable cue.

\subsection{Complementary structural features}
Beyond \texttt{centrality}, each cluster is described by the following label-free, session-relative features. We chose features that capture the teacher's \emph{behavioral} signature in a lesson (speaking the most, throughout the session, from the acoustic center of the room) rather than any voice-identity or acoustic cue tied to a specific speaker. Because every feature is session-relative and label-free, the same set transfers across rooms, speakers, and both languages, letting a single learned rule generalize without per-recording tuning.
\begin{itemize}
  \item \texttt{talk\_time\_frac}: share of total speech time held by the cluster. Computed on VAD-cleaned window durations (not raw transcript intervals) to avoid inflating teacher dominance.
  \item \texttt{coverage}: fraction of $20$ equal session-time bins in which the cluster appears. A teacher presents from bell to bell, scores 1; a student who speaks only in one group activity scores much lower.
  \item \texttt{spread}: temporal standard deviation of window midpoints, normalized by session length. It complements \texttt{coverage} since high coverage with low spread indicates a bursty cluster whose bins happen to be spread, while high spread confirms sustained presence.
  \item \texttt{turn\_count}: number of contiguous speaking runs in the time-ordered label sequence. Teachers typically hold many short turns (e.g., questions, transitions, corrections); students hold fewer.
  \item \texttt{size}: raw window count. It correlates with talk time, but is separately informative when window duration varies.
  \item \texttt{consistency}: mean cosine similarity of the cluster's windows to their own centroid $\mu_c$. A tight cluster (one voice) scores close to 1; a cluster polluted by mis-assigned voices scores lower.
  \item \texttt{energy\_z}: each window's log-RMS energy standardized within the recording (standard deviations above the session-mean level). It has a high score when teachers wear a microphone to record their lessons.
\end{itemize}

\paragraph{Why pooling across rooms is legitimate.}
To choose features, we pool every cluster from every recording, rank single features, and run greedy forward selection under leave-one-teacher-out cross-validation so that no teacher leaks between train and test. We optimize the \emph{worse} of the two languages, ensuring that the chosen set works for both languages. Pooling works even though the teacher is a different person each session because every feature is \emph{session-relative}. In other words, a teacher cluster lands in roughly the same region of feature space in every recording, whoever the teacher is. Pooling across hundreds of rooms, therefore, gives the classifier many labeled examples of that region, from which it learns a single decision boundary (``this region corresponds to the teacher'') that transfers frozen to a new room with no labels.

\subsection{Per-feature results}

\begin{figure}[t]
\centering
\includegraphics[width=0.995\linewidth]{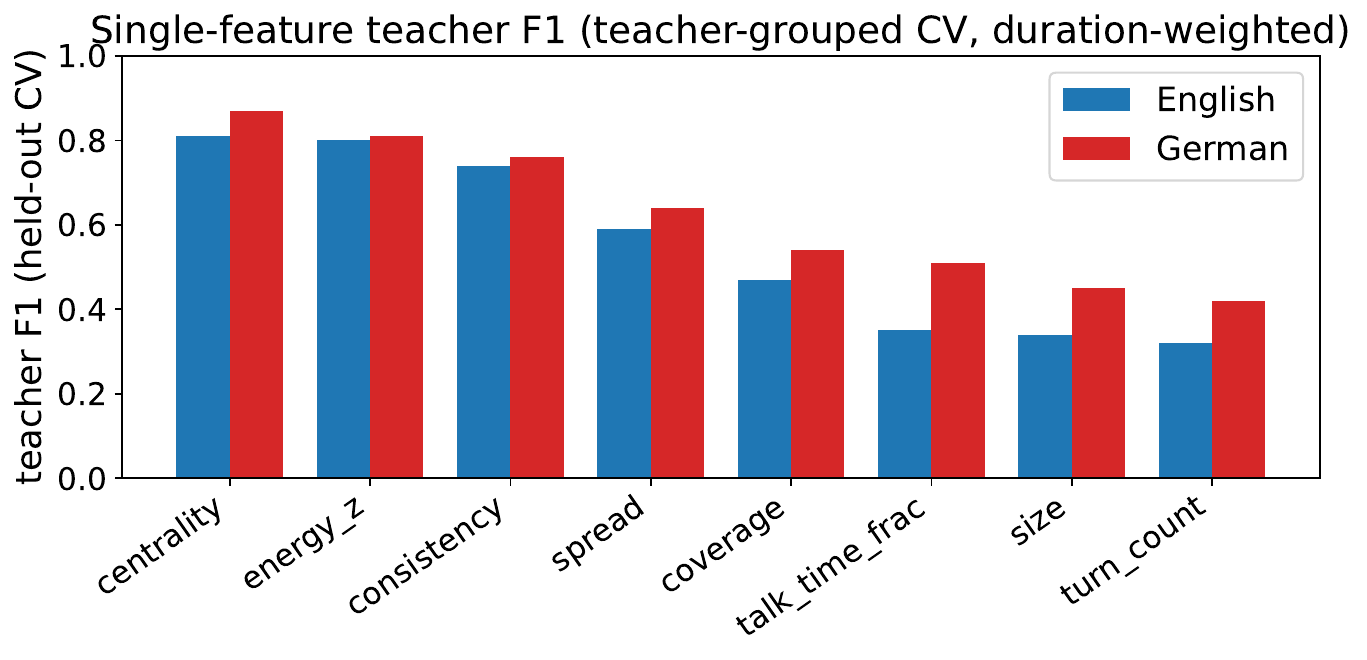}
\caption{Single-feature teacher F1. Centrality dominates and transfers across languages.}
\label{fig:features}
\end{figure}

To understand which acoustic and structural signals actually drive teacher identification, we evaluate each feature in isolation: we train the classifier on a single feature at a time, which is equivalent to reducing the teacher-identification step to that one cue alone. Figure~\ref{fig:features} shows the resulting teacher F1 per feature for both languages. The key finding is that the same compact feature set generalizes across both languages despite differences in recording setup and classroom culture, which is a necessary condition for a single deployable system.

\texttt{centrality} dominates across both languages (F1 $= 0.85$), confirming that the teacher's structural position in the turn-taking graph is the single most reliable cue: the teacher addresses every student, whereas students mostly address the teacher, making the teacher's centrality score a clear outlier among the clusters. Running greedy forward selection on top of this baseline adds almost no further gain, indicating that the remaining features are largely redundant once centrality is known.

\texttt{energy\_z}, the z-scored speaking energy relative to the session mean, is a useful second feature. Its contribution for German is mildly surprising given the far-field, reverberant recording condition; nonetheless, teachers in both corpora tend to speak with consistently higher energy than students, and that contrast survives acoustic degradation sufficiently to contribute marginal but consistent improvement.

The dominance-style features (\texttt{coverage}, \texttt{talk\_time\_frac}, and \texttt{turn\_count}) exhibit a characteristic high-precision, low-recall profile. They reliably identify the teacher when that speaker produces one large, contiguous dominant segment, but fail when teacher speech is fragmented across many short turns or when a verbose student temporarily matches the teacher's activity level. Because these features assign a high score to at most one cluster, they crown the right cluster but miss the scattered remainder of teacher speech, suppressing recall. 

\section{Session diagnostics: what predicts per-recording teacher F1}
\label{app:diag}

Section~\ref{sec:dominance} argues that per-recording accuracy is governed by
teacher/student separability. Here we substantiate that with a broader set of \emph{label-free} session descriptors. These quantities are computed at inference from the embeddings, the clustering, and the classifier's own outputs, using no ground-truth roles. For each of the $291$ recordings, we correlate every descriptor with the recording's teacher F1 (Spearman $\rho$).

\begin{table}[t]
\centering
\caption{Label-free session diagnostics vs.\ per-recording teacher F1
(Spearman $\rho$). Sorted by $|\rho|$. Descriptors above the rule track teacher
F1; those below do not.}
\label{tab:diag}
\small
\begin{tabular}{lc}
\toprule
diagnostic & $\rho$ \\
\midrule
classifier margin (mean)            & $0.82$ \\
separability (silhouette)           & $0.80$ \\
predicted teacher share $\hat{T}$   & $0.73$ \\
1-NN cluster-label agreement        & $0.73$ \\
teacher-cluster cohesion            & $0.70$ \\
global-mean$\to$teacher cosine      & $0.65$ \\
teacher-cluster fragmentation rate  & $-0.58$ \\
cluster-belief entropy              & $-0.55$ \\
teacher-duration Gini               & $0.48$ \\
\midrule
teacher/student centroid separation & $0.30$ \\
embedding effective rank            & $0.19$ \\
speech-vs-silence SNR proxy         & $0.19$ \\
centrality bimodality ($d'$)        & $-0.08$ \\
\bottomrule
\end{tabular}
\end{table}

\begin{figure}[t]\centering\includegraphics[width=.99\linewidth]{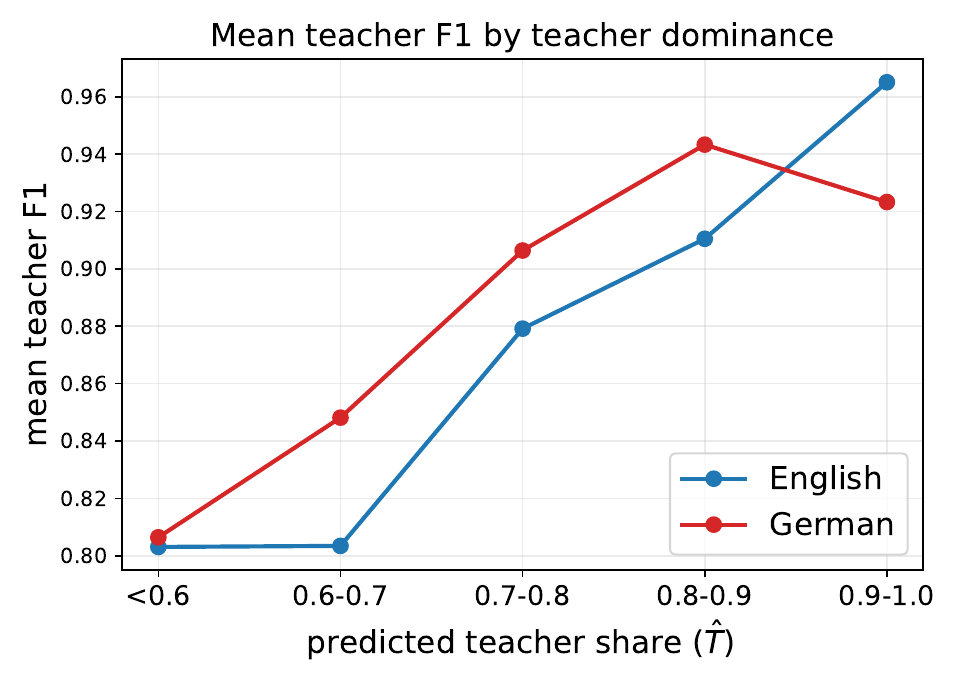}\caption{Mean per-recording teacher F1 with recordings binned by the model's predicted teacher share $\hat{T}$, which is a label-free quantity available at inference. Teacher F1 increases with $\hat{T}$ in both corpora, from about $0.80$ in the lowest bin to $0.92$--$0.97$ in the highest, so the low-scoring recordings concentrate at low $\hat{T}$ and can be flagged for human review without any labels.}\label{fig:predshare}\end{figure}

The strongest predictors, by a clear margin, are the classifier's own confidence and the separability of the split it produces. The mean \emph{classifier margin} (how far cluster scores sit from the decision boundary, 0.5) leads at $\rho = 0.82$, with the cosine \emph{silhouette} of the predicted teacher/student partition close behind ($0.80$). Both are label-free and both say the same thing: teacher F1 is high exactly when the two roles separate cleanly in the representation and the classifier is confident about which cluster is which, which is the mechanism Section~\ref{sec:dominance} identifies. A second group of geometry measures follows: 1-NN cluster-label agreement ($0.73$), teacher-cluster cohesion ($0.70$), the fragmentation of the teacher's speech across clusters ($-0.58$), and cluster-belief entropy ($-0.55$). Two dominance-flavoured descriptors also predict teacher F1: the predicted teacher share $\hat{T}$ ($0.73$) and the global-mean$\to$teacher cosine ($0.65$). This is expected, since a teacher who holds more of the floor is more fully sampled and easier to recover. These are useful, interpretable flags (we use $\hat{T}$ for exactly this in Figure~\ref{fig:predshare}), but they trail the separability and confidence measures rather than leading them.

Several intuitive descriptors carry little signal. Teacher/student centroid separation, despite being a separability measure, correlates only weakly ($\rho = 0.30$), as the silhouette already captures the same geometry more faithfully. Embedding \emph{effective rank} (a participation-ratio measure of anisotropy) was hypothesised to fall under reverberant channel collapse; it does not track teacher F1 ($\rho = 0.19$). Centrality bimodality is essentially noise ($-0.08$, wrong sign). Most notably, the \emph{acoustic} signal-to-noise ratio proxy, the ratio of mean frame log-energy in VAD-speech vs.\ VAD-silence regions, is only weakly correlated with F1 ($\rho = 0.19$), far below every geometry or confidence measure. This bears directly on the cross-language comparison in Section~\ref{sec:results}: the German corpus has the \emph{lower} SNR proxy ($8.9$ vs.\ $12.3$\,dB for English) yet the \emph{higher} F1, so the English$\leftrightarrow$German difference is structural (session composition and separability), not a matter of recording quality.

\section{Online cross-lesson accumulation}
\label{app:online}

\begin{figure}[t]
\centering
 \includegraphics[width=\linewidth]{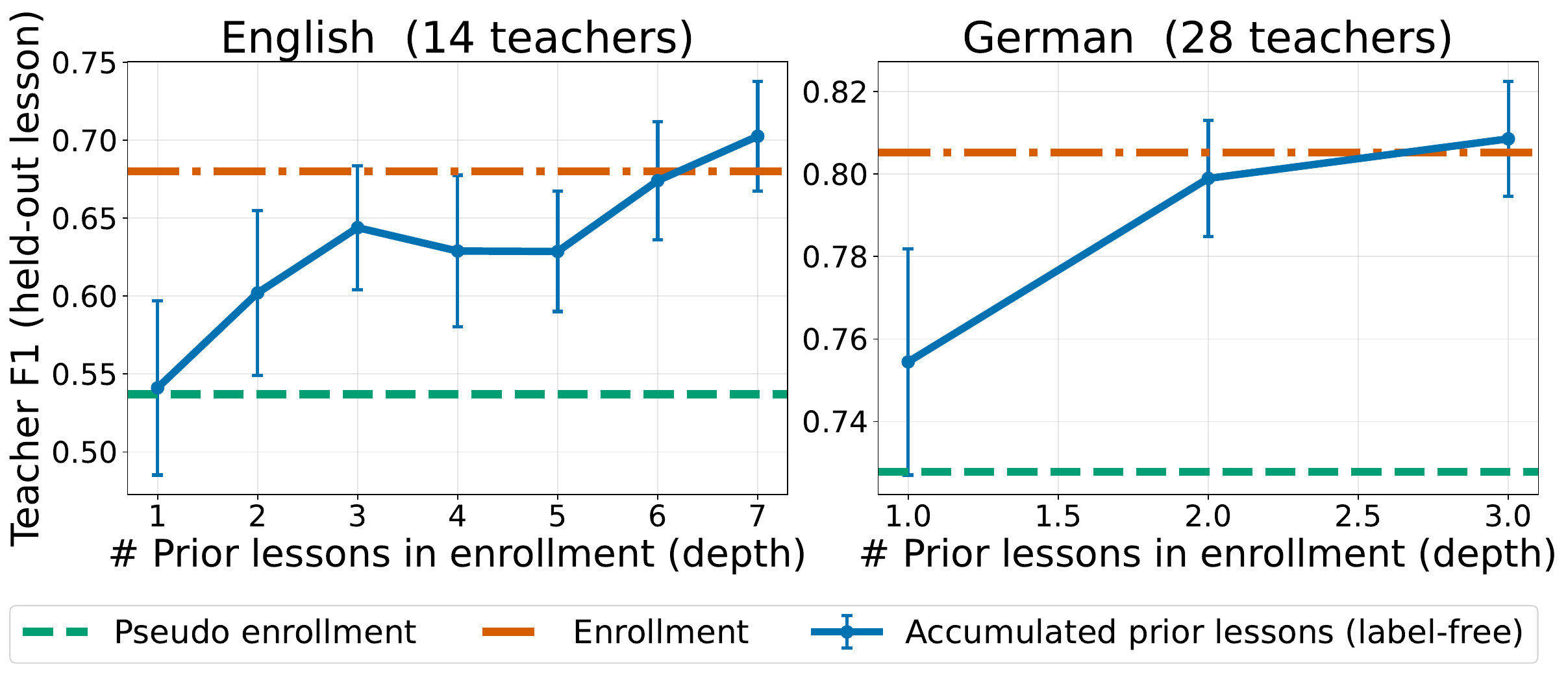}
\caption{Online cross-lesson teacher enrollment. Held-out teacher F1 vs.\ the number of prior lessons $k$ pooled into the teacher centroid. 
The blue curve pools only the teacher's prior lessons, selecting each lesson's teacher speech without labels. The green line is pseudo-enrollment (the label-free floor) and the orange line is the $30$-sec enrollment setting. Accumulation climbs from the pseudo-enrollment floor and saturates within two to three lessons, reaching the $30$-sec enrollment seed without ever using labels on the past lessons.}
\label{fig:online}
\end{figure}

The enrollment strategies of the main text treat each recording in isolation. But in the deployment we target, earlier recordings of the same teacher are available, which invites an \emph{online} variant: accumulate a teacher embedding from past lessons and use it to diarise the next one. We test this by grouping recordings by teacher and building the teacher centroid from the teacher's \emph{previous} lessons (we sweep the number of prior lessons $k$ and average over $30$ random lesson orderings per teacher). Figure~\ref{fig:online} plots held-out teacher F1 against $k$. It depicts two references: pseudo-enrollment (the fully label-free floor of Table~\ref{tab:strategies}) and the $30$-sec enrollment seed (which needs a hand-marked teacher clip). The accumulated centroid itself uses \emph{no labels} on any lesson, so the online curve is deployable end-to-end. Error bars are $\pm1$\,SE across teachers.

A single prior lesson already clears the label-free floor: at $k=1$ both languages sit above pseudo-enrollment, and accuracy then climbs monotonically as more history is pooled. German saturates within two to three lessons, reaching $0.81$ at $k=3$ and meeting the $30$-sec enrollment seed; English rises more gradually but overtakes the enrollment approach. The practical implication is the following: given a handful of a teacher's earlier recordings, label-free cross-lesson accumulation matches or exceeds the hand-seeded within-session enrollment. Hence, the manual enrollment clip and the labels behind it can be dropped entirely in the online deployment. 

We now ask whether our approach benefits from incorporating a teacher's previous lessons.  For this, we pool each teacher's predicted-teacher clusters across their earlier lessons into a single voiceprint and score every cluster by its cosine to it. This lifts German teacher F1 only from $0.87$ to $0.88$ and student F1 from $0.61$ to $0.63$ (no change in English). Interestingly, a \emph{single} prior lesson already yields this gain, with more history adding nothing. We attribute this to redundancy: across all clusters, a cluster's cosine to the pooled \textit{cross-lesson teacher voiceprint} and to the \textit{session-global mean embedding} track each other at $\rho = 0.97$ (still $0.89$ in the rooms with students speaking longer than teachers). In other words, a more accurate teacher voiceprint buys little: the teacher is recovered from discourse structure, not from the precision of a voiceprint.

\section{Masking mixed frames does not help}
\label{app:mask}

A natural suspect for the low-separability of some of the recordings is the \emph{mixed frames}: with $1.25$\,s windows at $0.5$\,s hop, any window may include a speaker change or genuine cross-talk. This would then produce a blended embedding that could drag cluster centroids off-target and corrupt the \texttt{centrality} feature. We tested this directly with a trained overlap detector (pyannote segmentation-3.0), which marks $8.7\%$ (English) / $3.9\%$ (German) of speech as overlapped, with the flagged windows over-representing students by $\sim$$1.7\times$, consistent with cross-talk over the teacher. Excluding every flagged window from the two embedding averages that feed \texttt{centrality} and \texttt{consistency} (leaving cluster assignment and all timing features untouched) moves teacher F1 by at most $0.002$ in either direction, across masking thresholds $0.1$--$0.5$ and across every low/mid/high separability stratum in both languages. This means a clean negative result, with no benefit even on the mixed-frame-heavy low-separability sessions that the intervention was meant to help. We believe that a duration-weighted mean over hundreds of clusters is barely perturbed by dropping $4$--$9\%$ of its members, so mixed-frame contamination never propagates through the centroid pathway far enough to matter. This reinforces Section~\ref{sec:dominance}: the tail is bounded by the \emph{separability} of the representation, not by frame-level label noise, so cleaning that noise upstream buys nothing and we keep the simpler pipeline without an overlap-detection stage.

\section{Deployability Test Implementation Details}
\label{app:bench}

\begin{table}[t]
\centering
\caption{Thread scaling of the proposed pipeline (recording~B, CPU execution provider):
RTF (median $\pm$ standard deviation over $N{=}3$ passes, lower is faster) as the CPU
thread budget varies.}
\label{tab:scaling}
\begin{tabular}{rrr}
\toprule
Threads & Phone & Laptop \\
\midrule
\ScalingBody
\bottomrule
\end{tabular}
\end{table}

\paragraph{Implementations.}
We benchmark the pipeline described in Sec.~\ref{sec:method}. The PyTorch implementation used for the accuracy experiments Silero VAD~\citep{silero2024}, the WeSpeaker \texttt{english\_resnet34} 256-d ResNet34-LM embedder~\citep{wang2023wespeaker}, agglomerative clustering at cosine cut $0.5$, and the role classifier which uses eight-feature session-relative. The Android implementation runs VAD and the exported embedder through ONNX Runtime, with clustering and role classification implemented natively. Both process one $1.25$\,s speech window per embedding call, with a $0.5$\,s hop and batch size one. The ONNX embedder returns an embedding of shape $[1,256]$. Its outputs have mean cosine similarity of $0.9993$ with the PyTorch outputs on held-out speech. We test the CPU, XNNPACK, and NNAPI execution providers on Android.

\paragraph{Baselines.}
The laptop baseline runs pyannote through \texttt{pyannote.audio}~3.1 in PyTorch, with evaluation mode, inference mode, and batch size one. The Android baseline uses \texttt{sherpa-onnx}~\citep{sherpaonnx} to run segmentation-3.0 and WeSpeaker through the same ONNX Runtime environment as our application. We report speedups separately for the laptop and phone because their baseline implementations differ.

\paragraph{Protocol.}
We measure real-time factor (RTF), peak resident set size (RSS), and processing time by pipeline stage. RTF is processing time divided by audio duration, so values below one indicate processing faster than real time. Model loading is timed separately and excluded from RTF. Each configuration receives one warm-up pass, followed by five timed passes for recording~A or three for recording~B. We report median RTF, with standard deviations in Table~\ref{tab:scaling} and minimum-to-maximum ranges in Fig.~\ref{fig:bench}

\paragraph{Hardware and data.}
We run the benchmarks on a 12-core Apple M4-Pro Macbook Pro laptop and a Samsung Galaxy~S24+(\texttt{SM-S926B}, 10-core \texttt{s5e9945} SoC, Android~16). Recordings~A and~B are two held-out $10$-minute clips from the German GTI corpus, recorded with room microphones and processed as $16$\,kHz mono audio. The primary runtime comparison uses one inference thread. We separately test thread counts of $\{1,2,4,8\}$ on both devices. On the phone, we also compare execution providers at one thread, obtaining RTFs of $0.228$ for CPU, $0.221$ for XNNPACK, and $0.263$ for NNAPI.

\paragraph{Per-stage cost.}
Speaker embeding accounts for approximately $96\%$ of processing time on the phone and $91\%$ on the laptop. VAD accounts for $3$-$9\%$, while clustering and role assignment together account for less than $2\%$. The pyannote baseline takes approximately $507$\,s per pass on the phone. We report its total runtime without a per-stage breakdown.

\section{Downstream transcription analysis}
\label{app:wer}

\paragraph{Protocol.}
All $292$ recordings are transcribed with Whisper-large-v2 in fixed $30$\,s chunks whose returned text is concatenated, under two conditions: \emph{no filtering} (the whole recording) and \emph{pipeline} (only the spans the classifier labels teacher, at the deployed operating point $\tau{=}0.45$). Both are scored against the teacher's turns in the human transcript, and \emph{only teacher words are counted}: student speech is neither credited when recognised nor charged when transcribed from unfiltered audio, so the comparison isolates recognition of the teacher's own words rather than rewarding the removal of student speech.

\begin{table}[h]
\centering
\caption{Teacher word error rate with and without role filtering, pooled over reference
words within each corpus. Substitutions and insertions fall under filtering in both
corpora; deletions rise and dominate, because an imperfect filter mutes some genuine
teacher speech that the recogniser then never sees. On German the two conditions are
statistically indistinguishable (macro-averaged over recordings: $0.3635$ without filtering
versus $0.3631$ with; $69$ recordings improve and $71$ worsen).}
\label{tab:wer}
\setlength{\tabcolsep}{4pt}
\begin{tabular}{llrrrr r}
\toprule
Corpus & Condition & WER & Sub & Del & Ins & Kept \\
\midrule
English & no filtering & \textbf{0.160} & 0.059 & 0.040 & 0.061 & 100\% \\
        & pipeline     & 0.213 & 0.049 & \textbf{0.135} & 0.029 & 52\% \\
\midrule
German  & no filtering & 0.361 & 0.144 & 0.145 & 0.073 & 100\% \\
        & pipeline     & \textbf{0.352} & 0.124 & \textbf{0.192} & 0.036 & 40\% \\
\bottomrule
\end{tabular}
\end{table}

\begin{figure*}[t]
\centering
\includegraphics[width=0.98\linewidth]{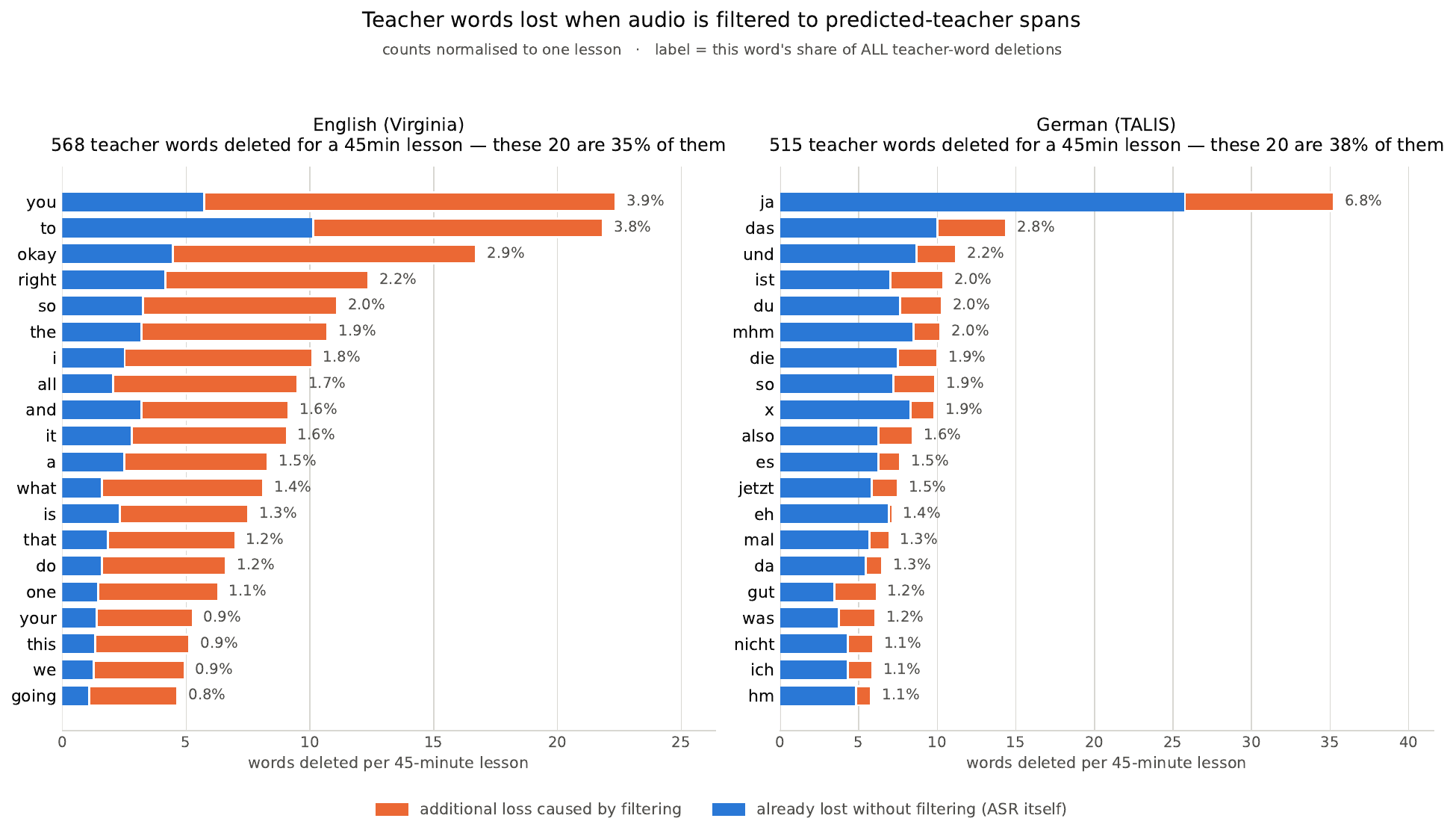}
\caption{The teacher words most often lost under role filtering (top $20$ per corpus, counts normalised to one $45$-minute lesson). Each bar is split into two: the lower segment is celetions already present \emph{without} filtering, i.e.\ imperfect transcription because of the recorded audio, and the upper segment the additional loss filtering causes. The distinction matters, because Whisper already omits many German fillers from untouched audio (\emph{eh} is $77\%$ deleted
with no filtering at all). The lost vocabulary is dominated by fillers, backchannels and discourse markers in both languages.}
\label{fig:delwords}
\end{figure*}

\end{document}

%% file: tables.tex
\newcommand{\StrategiesBody}{%
  \multirow{4}{*}{\shortstack[l]{English\\(24 teachers)}} & pseudo\_enroll & $0.79\pm0.12$ & $0.43\pm0.20$ & $0.53\pm0.20$ & $0.43\pm0.09$ & $\mathbf{0.80\pm0.10}$\\
   & enrollment & $0.83\pm0.03$ & $0.55\pm0.15$ & $0.65\pm0.12$ & $0.47\pm0.07$ & $0.75\pm0.11$\\
   & our pipeline & $0.76\pm0.06$ & $\mathbf{0.87\pm0.04}$ & $0.81\pm0.04$ & $\mathbf{0.64\pm0.07}$ & $0.46\pm0.11$\\
   & \textit{pyannote-diar}$^\dagger$ & $\mathbf{0.86\pm0.06}$ & $0.83\pm0.03$ & $\mathbf{0.84\pm0.08}$ & $0.40\pm0.04$ & $0.58\pm0.05$\\
  \midrule
  \multirow{4}{*}{\shortstack[l]{German\\(39 teachers)}} & pseudo\_enroll & $0.90\pm0.04$ & $0.55\pm0.22$ & $0.66\pm0.20$ & $0.48\pm0.12$ & $\mathbf{0.87\pm0.06}$\\
   & enrollment & $\mathbf{0.92\pm0.03}$ & $0.67\pm0.14$ & $0.77\pm0.10$ & $0.55\pm0.09$ & $0.85\pm0.07$\\
   & our pipeline & $0.83\pm0.08$ & $\mathbf{0.91\pm0.04}$ & $\mathbf{0.87\pm0.05}$ & $\mathbf{0.76\pm0.06}$ & $0.60\pm0.10$\\
   & \textit{pyannote-diar}$^\dagger$ & $0.84\pm0.10$ & $0.80\pm0.09$ & $0.82\pm0.09$ & $0.53\pm0.05$ & $0.74\pm0.07$\\
}

\newcommand{\ScalingBody}{%
$1$ & $0.228{\pm}0.001$ & $0.027{\pm}0.0001$ \\
$2$ & $0.190{\pm}0.008$ & $0.031{\pm}0.0002$ \\
$4$ & $0.137{\pm}0.005$ & $0.027{\pm}0.00002$ \\
$8$ & $\mathbf{0.133{\pm}0.008}$ & $0.036{\pm}0.0004$ \\
}

\newcommand{\EmbedClfBody}{%
\multirow{2}{*}{English} & ResNet34  & $0.76\pm0.06$ & $0.87\pm0.04$ & $0.81\pm0.04$ & $0.64\pm0.07$ & $0.46\pm0.11$\\
 & ResNet221 & $0.77\pm0.05$ & $0.87\pm0.04$ & $0.81\pm0.04$ & $0.64\pm0.06$ & $0.48\pm0.11$\\
\midrule
\multirow{2}{*}{German} & ResNet34  & $0.83\pm0.08$ & $0.91\pm0.04$ & $0.87\pm0.05$ & $0.76\pm0.06$ & $0.60\pm0.10$\\
 & ResNet221 & $0.83\pm0.08$ & $0.92\pm0.04$ & $0.87\pm0.05$ & $0.77\pm0.06$ & $0.60\pm0.10$\\
\midrule
\multirow{2}{*}{Joint} & ResNet34  & $0.81\pm0.08$ & $0.89\pm0.04$ & $0.85\pm0.06$ & $0.71\pm0.08$ & $0.55\pm0.13$\\
 & ResNet221 & $0.81\pm0.08$ & $0.90\pm0.05$ & $0.85\pm0.06$ & $0.72\pm0.08$ & $0.55\pm0.12$\\
}

\newcommand{\EmbedEnrollBody}{%
\multirow{2}{*}{English} & ResNet34  & $0.83\pm0.03$ & $0.55\pm0.15$ & $0.65\pm0.12$ & $0.47\pm0.07$ & $0.75\pm0.11$\\
 & ResNet221 & $0.82\pm0.03$ & $0.60\pm0.15$ & $0.68\pm0.11$ & $0.49\pm0.07$ & $0.72\pm0.12$\\
\midrule
\multirow{2}{*}{German} & ResNet34  & $0.92\pm0.03$ & $0.67\pm0.14$ & $0.77\pm0.10$ & $0.55\pm0.09$ & $0.85\pm0.07$\\
 & ResNet221 & $0.91\pm0.03$ & $0.71\pm0.13$ & $0.79\pm0.10$ & $0.57\pm0.09$ & $0.84\pm0.07$\\
\midrule
\multirow{2}{*}{Joint} & ResNet34  & $0.88\pm0.05$ & $0.63\pm0.15$ & $0.72\pm0.12$ & $0.52\pm0.09$ & $0.81\pm0.10$\\
 & ResNet221 & $0.88\pm0.05$ & $0.67\pm0.15$ & $0.75\pm0.12$ & $0.54\pm0.09$ & $0.80\pm0.11$\\
}